\documentclass{article}

\usepackage[final]{corl_2020} 

\usepackage{amsmath}
\usepackage{amssymb}
\usepackage{booktabs}
\usepackage{glossaries}
\usepackage{float}
\usepackage{graphicx}
\usepackage{lipsum}
\usepackage{multirow}
\usepackage{siunitx}
\usepackage{wrapfig}

\glsdisablehyper  

\renewcommand{\vec}[1]{\mathbf{#1}}

\newacronym{dof}{DOF}{Degrees of Freedom}
\newacronym{fcn}{FCN}{Fully Convolutional Network}
\newacronym{gpd}{GPD}{Grasp Pose Detection}
\newacronym{tsdf}{TSDF}{Truncated Signed Distance Function}
\newacronym{vgn}{VGN}{Volumetric Grasping Network}

\title{Volumetric Grasping Network: Real-time\\6 DOF Grasp Detection in Clutter}

\author{
  Michel Breyer \\
  ETH Z\"urich \\
  \texttt{mbreyer@ethz.ch} \\
  \And
  Jen Jen Chung \\
  ETH Z\"urich \\
  \texttt{chungj@ethz.ch} \\
  \And
  Lionel Ott \\
  University of Sydney \\
  \texttt{lionel.ott@sydney.edu.au} \\
  \AND
  Roland Siegwart \\
  ETH Z\"urich \\
  \texttt{rsiegwart@ethz.ch} \\
  \And
  Juan Nieto \\
  ETH Z\"urich \\
  \texttt{nietoj@ethz.ch} \\
}

\begin{document}
\maketitle


\begin{abstract}
General robot grasping in clutter requires the ability to synthesize grasps that work for previously unseen objects and that are also robust to physical interactions, such as collisions with other objects in the scene. In this work, we design and train a network that predicts 6 DOF grasps from 3D scene information gathered from an on-board sensor such as a wrist-mounted depth camera. Our proposed \gls{vgn} accepts a \gls{tsdf} representation of the scene and directly outputs the predicted grasp quality and the associated gripper orientation and opening width for each voxel in the queried 3D volume. We show that our approach can plan grasps in only \SI{10}{ms} and is able to clear 92\% of the objects in real-world clutter removal experiments without the need for explicit collision checking. The real-time capability opens up the possibility for closed-loop grasp planning, allowing robots to handle disturbances, recover from errors and provide increased robustness. Code is available at \url{https://github.com/ethz-asl/vgn}.
\end{abstract}



\section{Introduction} \label{sec:introduction}

Traditionally robotic manipulation has mostly considered repetitive tasks performed in tightly controlled spaces. However, recently there has been a lot of interest for deploying robots to domains that require more flexibility. For example, an assistant robot could alleviate the workload of medical staff by taking over the time-consuming task of fetching supplies in a hospital. To perform well in such unstructured environments, the system must be able to compute grasps for the sheer unlimited number of objects it might encounter, and at the same time deal with clutter, occlusions, and high-dimensional noisy readings from on-board sensors.

Due to these challenges, recent research in grasp synthesis has overwhelmingly favored data-driven approaches to plan grasps directly from sensor data, outperforming manually designed policies~\citep{bohg2014datadriven,lenz2015deep,mahler2017dexnet}. However, the planning is usually constrained to top-down grasps from single depth images. This type of approach enables compact representations by constraining the search to grasps perpendicular to the image plane, assuming therefore a favorable camera placement and restricting the robot to grasp objects from a single direction. This limits the flexibility of the system, as in some cases it is easier to approach different objects in the scene from different directions~\cite{gualtieri2016high,lundell2020beyond}. Several notable recent works tackle full 6 \gls{dof} grasp pose detection~\cite{gualtieri2016high,choi2018learning,mousavian20196dof,murali20206dof}. However these methods often reason on single, isolated objects requiring additional collision checks when placed within the scene, or their rather high computation times (in the order of seconds) make them unsuitable for closed-loop execution, which is essential for more advanced interaction and reactivity to dynamic changes.

\begin{figure}
    \centering
    \includegraphics[width=\textwidth]{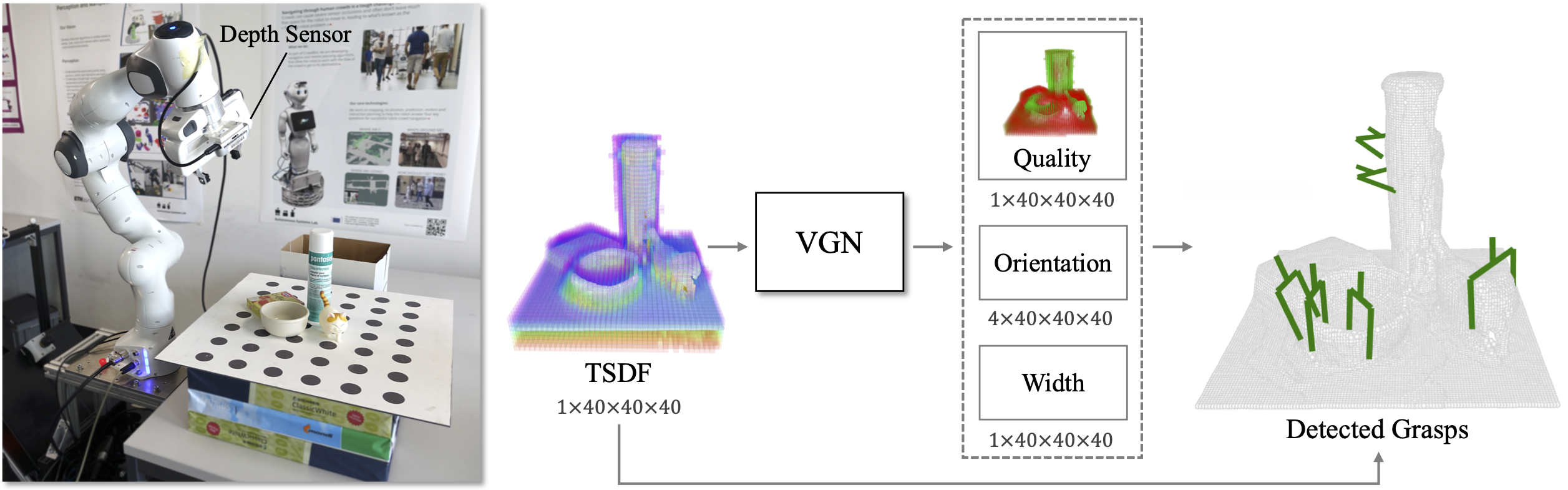}
    \caption{A scan of the 3D scene is converted into the \gls{tsdf} that is passed into \gls{vgn}. \gls{vgn} has a three-headed output, providing the grasp quality, gripper orientation and gripper width at each voxel. Non-maxima suppression is then applied using the grasp quality output and invalid grasps are filtered out given the input \gls{tsdf}.}
    \label{fig:overview}
\end{figure}

In this work, we propose a novel approach to real-time 6 \gls{dof} grasp synthesis. The input to our algorithm is a 3D voxel grid, where each cell contains the truncated distance to the nearest surface. \glspl{tsdf} not only offer an elegant framework to fuse multiple observations into a consistent map while smoothing out sensor noise, their regular structure also makes them suitable for feature learning via deep neural networks. We train a \gls{fcn} to map the input \gls{tsdf} to a volume of the same spatial resolution, where each cell contains the predicted quality, orientation, and width of a grasp executed at the center of the voxel (see Figure~\ref{fig:overview}). The network is trained on a synthetic dataset of cluttered grasp trials generated using physical simulation. The integrated nature of our approach allows us to detect grasps for the whole workspace with a single forward pass of the network. We based our approach on the hypothesis that the inclusion of 3D information of the full scene allows the neural network to capture collisions between the gripper and its environment, which is essential in clutter. 
We validate our hypothesis with a series of simulated experiments. The results show competitive grasping performance, while only requiring \SI{10}{ms} on a GPU equipped computer. We also run 80 grasping trials on a physical setup, demonstrating that the approach transfers to a real setup without any additional fine-tuning.


In summary, the contribution of this work is a novel grasping network that:

\begin{itemize}
    \item Enables 6 \gls{dof} grasp synthesis in real-time, and
    \item Uses the full 3D scene information to directly learn collision-free grasp proposals.
\end{itemize}

\section{Related Work} \label{sec:related-work}

Robotic grasping has been a topic of interest for several decades~\cite{bohg2014datadriven}. Recent contributions in this field have overwhelmingly favored the use of deep learning approaches for detecting robot grasps directly from sensor data. This bias is not unwarranted as deep learning methods for robot grasping have demonstrated superior performance in terms of generalizing to previously unseen objects and finding successful grasps in cluttered scenes~\cite{mahler2017dexnet,gualtieri2016high}.

Grasp synthesis approaches can broadly be split into those that output ``top-down'' grasps (typically 3 or 4 \gls{dof}) and those that can output 6 \gls{dof} grasps. Each category is tightly coupled with the scene representation used as input to the learned network. Top-down grasps either take overhead visual data~\cite{pinto2016supersizing}, depth image data~\cite{mahler2017dexnet, johns2016deep, morrison2018closing}, or a combination of both~\cite{lenz2013deep, kumra2017robotic} and return viable grasps in the image frame. In comparison, networks that synthesize 6 \gls{dof} grasps take in the full 3D information of the scene, which typically comes in the form of a point cloud~\cite{gualtieri2016high, mousavian20196dof, liang2019pointnetgpd} or occupancy grid~\cite{choi2018learning,lu2019modeling}. While point clouds are able to identify object surfaces in space, we believe that the additional distance-to-surface information provided by a \gls{tsdf} representation~\cite{curless1996volumetric} can improve overall grasp detection performance. Distance functions are already commonly used in collision-free path planning~\cite{oleynikova2017voxblox}, and have been applied to enforce collision-free grasp configurations for a multi-fingered hand~\cite{vandermerwe2020learning}. Furthermore, in line with the conclusions of \cite{lundell2020beyond,murali20206dof}, we also expect grasp performance to improve when the full scene is considered during grasp synthesis (as opposed to, for example, only considering a subset of the point cloud~\cite{gualtieri2016high}) since it allows the system to account for physical interactions such as collisions with other objects. In our work, we provide the full \gls{tsdf} to the network, thereby avoiding the need for separate processing steps to handle scene completion \cite{lundell2020beyond} or explicit collision detection \cite{murali20206dof}.

It is worth noting that several of these grasp synthesis methods primarily act as grasp quality predictors, meaning that they still require some initial process to sample grasps to evaluate. For example, the Dex-Net 2.0 grasp planner~\cite{mahler2017dexnet} initially computes antipodal grasps from the input depth image, while \citet{gualtieri2016high} uniformly sample several thousand grasp candidates over the object point cloud and return only those predicted to result in successful grasps. PointNetGPD~\cite{liang2019pointnetgpd} improves on this by providing a grasp score that is trained on grasp wrench space analysis and force closure metrics. Nevertheless, requiring an initial sampling step can be problematic since it forces a trade off between computational tractability and ensuring adequate sampling coverage over the grasp space. GraspNet~\cite{mousavian20196dof} addresses this by training a variational grasp sampler that produces promising grasp candidates given the input point cloud. The sampled grasps are then iteratively refined using a grasp pose evaluator, which is a process somewhat similar to the analysis-by-synthesis procedure of~\cite{yan2018learning}.

In our work, we take inspiration from the 2D grasp map of~\cite{morrison2018closing, zeng2018learning, satish2019onpolicy} and train our grasp network to directly output a grasp (orientation and gripper width) and associated grasp quality prediction for each voxel in the entire volumetric representation of the scene. The network output can then be directly queried for the highest scoring grasp or set of grasps using non-local maxima suppression. This avoids having to sample candidate grasps at run-time or iterate over grasp proposals to find viable grasps, both of which can be computationally expensive.


\section{Problem Formulation} \label{sec:problem-formulation}

We consider the problem of planning parallel-jaw grasps for unknown rigid objects in clutter. The goal is to find gripper configurations that allow the robot to successfully grasp and remove the objects from the workspace. Instead of sampling and evaluating individual candidates, we want to evaluate a large number of discretized grasp positions in parallel.

The visual input is represented as a \gls{tsdf},  an $N^3$ voxel grid $\mathbf{V}$ where each cell $V_{i}$ contains the truncated signed distance to the nearest surface. The edges of the volume correspond to the boundaries of the grasping workspace of size $l$.
We define a grasp $g$ by the position $\vec{t}$ and orientation $\mathbf{r}$ of the gripper with respect to the robot's base coordinate system, as well as the opening width $w$ of the gripper. In addition, each pose is associated with a scalar quantity $q \in [0,1]$ capturing the probability of grasp success. Given the voxel size $v$ and the rigid transformation $T_{RV}$ between the base and the \gls{tsdf} volume frames, we can describe a grasp~$\tilde{g}$ by

\begin{equation} \label{eq:transformation}
    \tilde{\vec{t}} = T_{RV}(\vec{t}) / v, \tilde{\vec{r}} = T_{RV}(\vec{r}), \tilde{w} = w / v,
\end{equation}

where $\tilde{\vec{t}}$ corresponds to the index of the voxel at which the grasp is defined and the width $\tilde{w}$ is expressed in terms of voxel size units. Our goal is to find a map $f: \vec{V} \to \vec{Q}, \vec{R}, \vec{W}$, where $\vec{Q}$, $\vec{R}$, and $\vec{W}$ contain the quality $q$, orientation $\tilde{\vec{r}}$, and opening width $\tilde{w}$ of a grasp at each voxel $\tilde{\vec{t}}$. While regressing to a single orientation cannot capture the full distribution of grasps at a given voxel, we think that a point estimate is enough to predict a good grasp while reducing the complexity of the approach. The output is a dense grasp map that can be further filtered based on task-specific constraints for a final grasp selection, as explained in more detail in Section~\ref{sec:detection}.


\section{Volumetric Grasping Network}\label{sec:vgn}

In this section, we describe the \acrfull{vgn}, a deep neural network approximation of the dense grasp map $f$ defined in the previous section.

\subsection{Architecture}

Our network follows a \gls{fcn} architecture. First, a \emph{perception} module consisting of 3 strided convolutional layers with $16, 32$ and $64$ filters, respectively, maps the input volume $\mathbf{V}$ to a feature map of dimension $64 \times 5^3$. The second part of the network consists of 3 convolutional layers interleaved with $2\times$ bilinear upsampling, followed by three separate heads for predicting grasp quality, rotation, and opening width. The \emph{grasp quality} head outputs a volume of size $1 \times N^3$, where every entry $\hat{q}_i$ represents the predicted probability of success of a grasp executed at the center of the voxel. The \emph{rotation} head regresses to a quaternion representation of the orientation of the associated grasp candidates. We chose quaternions over alternative representations, such as Euler angles, as they were found to perform better in learning forward kinematics~\cite{grassmann2019merits}. A normalization layer at the end of the rotation branch ensures unit quaternions. Finally, the \emph{width} head predicts the opening width of the gripper at each voxel. 

\subsection{Synthetic Training} \label{sec:synthetic-training}

The full network is trained end-to-end on ground-truth grasps obtained by simulated trial using the the following loss function,

\begin{equation}
    \mathcal{L}(\hat{g}_i, \tilde{g}_i) = \mathcal{L}_q(\hat{q}_i, q_i) + q_i \left(\mathcal{L}_r(\hat{\vec{r}}_i, \tilde{\vec{r}}_i) +  \mathcal{L}_w(\hat{w}_i,  \tilde{w}_i)\right),
\end{equation}

where $q_i \in \{0,1\}$ denotes the ground truth grasp label of a target grasp $\tilde{g}_i$. $\mathcal{L}_q$ is the binary cross-entropy loss between predicted and ground-truth labels $\hat{q}$ and $q$, and $\mathcal{L}_w$ is the mean-squared error between predicted and target widths $\hat{w}$ and $\tilde{w}$, respectively. We use the inner product to compute the distance between the predicted and target quaternions, $\mathcal{L}_{quat} = 1 - |\hat{\vec{r}} \cdot \tilde{\vec{r}}|$. This loss formulation is derived from the angle formed by the two quaternions~\citep{kuffner2004effective}. However, it does not handle the symmetry of a parallel-jaw gripper. A configuration rotated by \SI{180}{\degree} around the gripper's wrist axis corresponds effectively to the same grasp, but leads to inconsistent loss signals as the network gets penalized for regressing to one of the two alternative 3D rotations. To avoid this problem, we extend the loss function to consider both correct rotations as ground truth options,

\begin{equation}
    \mathcal{L}_r(\hat{\vec{r}}, \tilde{\vec{r}}) = \min\left( \mathcal{L}_{quat}(\hat{\vec{r}}, \tilde{\vec{r}}), \mathcal{L}_{quat}(\hat{\vec{r}}, \tilde{\vec{r}}_\pi) \right).
\end{equation}

\begin{figure}[t]
    \centering
    \includegraphics[width=\textwidth]{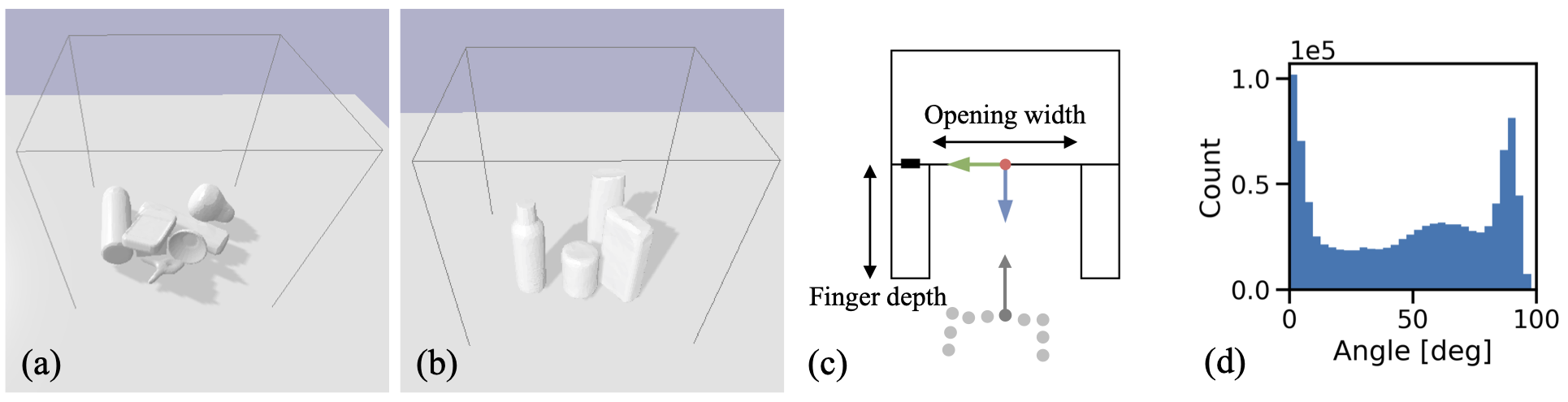}
    \caption{Examples of ``pile'' and ``packed'' scenes (a) and (b) respectively. Subfigure (c) shows the definition of the grasp frame origin with respect to the gripper geometry and (d) shows the distribution of angles between the gravity vector and the $z$ axis of grasps from the training set.}
    \label{fig:data_generation}
\end{figure}

Since it is impractical to collect target values for every voxel of the network output, we only compute the loss and gradients through the voxels at which ground-truth data are available. In order to generate a diverse set of labeled grasps spanning different amounts of clutter and the whole configuration space of the gripper, we created virtual scenes for simulated grasping trials using two different strategies, ``pile'' and ``packed''. In the first scenario, objects are dropped into a box placed on a flat surface. Removing the container leaves a cluttered pile of objects. In the second scenario, a subset of taller objects are iteratively placed upright at random locations in the workspace, rejecting positions that end up in collision with already placed objects. The first scenario favors top-down grasps while the second scenario encourages side-grasps. In both cases, in each new scene we spawn $m$ objects, where $m \sim \mathrm{Pois(4)} + 1$, though the placing procedure of ``packed'' scenes is interrupted after a given maximum number of attempts. An example for each scenario is shown in Figure~\ref{fig:data_generation}.

After the virtual scene has been created, we reconstruct a point cloud by fusing $n \sim \mathcal{U}(1,6)$ synthetic depth images. The viewpoints of the virtual camera pointing towards the center of the workspace with length $l$ are sampled using spherical coordinates $r \sim \mathcal{U}(1.6l, 2.4 l)$, $\theta \sim \mathcal{U}(0, \frac{\pi}{4})$, and $\phi \sim \mathcal{U}(0, 2 \pi)$. Next, we sample a point and associated normal from the reconstructed cloud. Applying a random offset along the normal yields the position of the sample grasp candidate. In order to obtain a ground truth grasp label, we test 6 different grasp orientations about the normal vector by spawning the gripper at the given pose, then attempting to close the jaws and retrieve the object. Configurations that lead to a collision with the environment are labeled as negative. In case of a positive outcome, we store the orientation and gripper width in addition to the position and label. In order to reuse storage and computation, we sample and evaluate 120 points from each cloud.

Using the described procedure, we generate a balanced dataset of approximately 2 million grasps by discarding superfluous negative samples. Figure~\ref{fig:data_generation}(d) shows the distribution of the angle between the gravity vector and the $z$ axis of the grasp over the training set. We can see that both top-down ($\sim$\SI{0}{\degree}) and side grasps ($\sim$\SI{90}{\degree}) are well represented. Finally we train the \gls{fcn} with the Adam optimizer and a learning rate of $3\times10^{-4}$ for 10 epochs with batch sizes of 32. Samples are augmented with random rotations of multiples of \SI{90}{\degree} about the gravity vector and random offsets along the $z$ axis.

\subsection{Grasp Detection}\label{sec:detection}

Given the output volumes of a trained \gls{vgn}, we perform several steps in order to compute a list of promising grasp candidates for a given scene. First, the grasp quality tensor is smoothed with a 3D Gaussian kernel which favors grasps in regions of high grasp quality~\cite{johns2016deep}. Next, we mask out voxels whose distance to the nearest surface is smaller than the finger depth as defined in Figure~\ref{fig:data_generation}(c). This step is required since we do not include such configurations in the training data as they cannot possibly lead to successful grasps. A threshold on the grasp quality masks out voxels with low predicted scores and we apply non-maxima suppression. We construct a list of grasp candidates from the remaining voxel indices by looking up the values for orientation and width from the respective tensors. These grasp candidates are then transformed back into Cartesian coordinates using the inverse of Equation~\ref{eq:transformation}. Further filters could be applied depending on task-specific needs, e.g. enforcing some constraints on the approach vector. An overview of the whole pipeline is shown in Figure~\ref{fig:overview}.


\section{Experiments}

We perform a series of simulated and real robot experiments to evaluate our grasp detection pipeline. The goals of the experiments are to (a) investigate the grasping performance of the approach, (b) verify real-time grasp detection rates, and (c) determine whether the network can be transferred to a physical system without additional fine-tuning.

\subsection{Experimental Setup}

\begin{wrapfigure}[15]{R}{0.45\textwidth}
    \centering\vspace{-0.5cm}
    \includegraphics[width=0.45\textwidth]{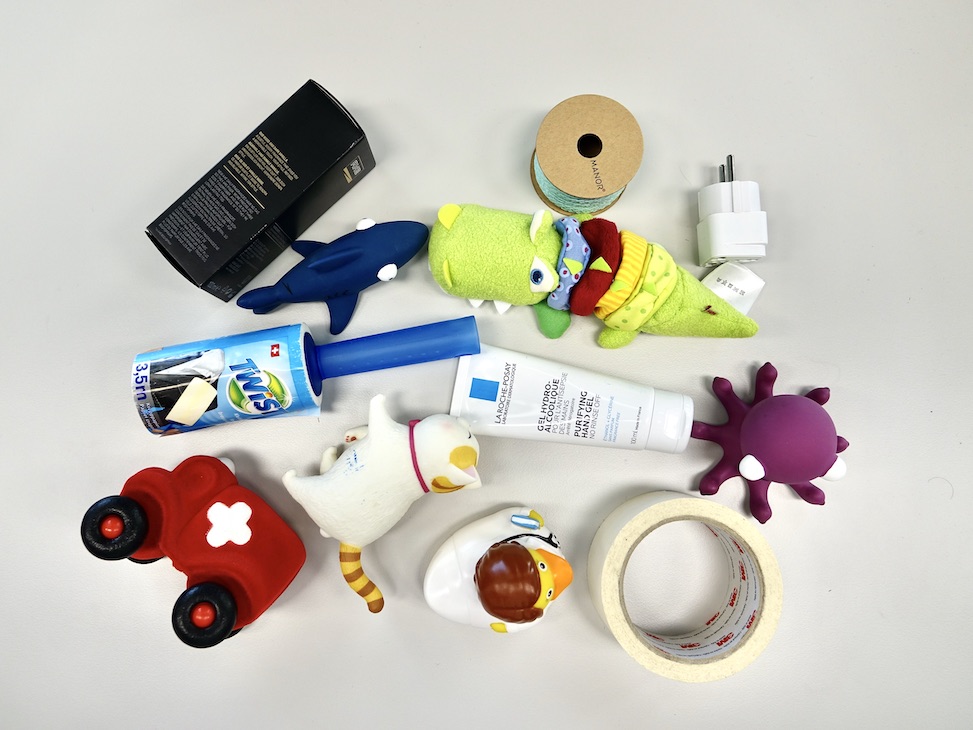}\vspace{-0.5cm}
    \caption{The 12 test objects used in our robot grasping experiments.}
    \label{fig:object_set}
\end{wrapfigure}

The experimental setup consists of a 7 \gls{dof} Panda arm from Franka Emika with a stock parallel-jaw gripper. Grasps are planned in a $30 \times 30 \times \SI{30}{cm^3}$ workspace located in front of the robot. For perception, we mounted an Intel RealSense D435 depth sensor to the wrist of the hand as shown in Figure~\ref{fig:overview}. The transformation between the gripper and camera is calibrated using the procedure from~\citet{furrer2018evaluation} and the $640 \times 480$ depth images are integrated into a \gls{tsdf} with a resolution of $N = 40$ using the implementation from Open3D~\citep{zhou2018open3d}. ROS is used to handle all communication with the robot and the sensor. We collected a set of 12 test objects (shown in~Figure~\ref{fig:object_set}) from classes similar to related works~\cite{gualtieri2016high, morrison2018closing}. We did not consider any transparent, or reflective objects due to the limitations of the depth sensor.

We also built a simulated environment in PyBullet~\cite{coumans2020pybullet} mimicking the real setup which allows us to generate large data sets of grasp trials, and to perform more extensive evaluations. Synthetic depth images are generated using the bundled software renderer. We assembled a set of 343 object meshes from different sources~\citep{kasper2012kit, singh2014bigbird, calli2015ycb, kappler2015leveraging} and split them into 303 training and 40 testing objects. All meshes were inspected for water tightness, simplified, and scaled such that at least one side fits within the gripper. The whole pipeline is implemented in Python. Network training and simulated experiments were performed on a computer equipped with an Intel Core i7-8700K and a GeForce GTX 1080 Ti graphics card, while all computations on the robotic platform were performed on a CPU-only machine with an Intel Core i7-8550U.

\subsection{Simulated Experiments}

\begin{table}[t]
    \centering
    \caption{Success rates (\%) and \% cleared for simulated picking experiments on the two scenarios with object counts $m = \{5, 10\}$ averaged over 200/100 rounds respectively.}
    \begin{tabular}{l c c c c c}
        \toprule
        Method & \multicolumn{3}{c}{5 objects} & \multicolumn{2}{c}{10 objects} \\
        \cmidrule(lr){2-4}
        \cmidrule(lr){5-6}
        & Blocks & Pile & Packed & Blocks & Pile \\ 
        \midrule
        GPD  & 88.6 / 39.4 & 59.9 / 26.1 & 73.7 / 72.8 & \textbf{87.7} / 24.8 & \textbf{63.1} / 17.0 \\
        VGN ($\epsilon = 0.95$) & \textbf{89.5} / 85.9 & \textbf{65.4} / 41.6 & \textbf{91.5} / 79.0 & 85.3 / 66.7 & 59.4 / 25.1 \\
        VGN ($\epsilon = 0.90$) & 87.6 / \textbf{90.1} & 62.3 / 46.4 & 87.6 / \textbf{80.4} & 82.5 / \textbf{77.6} & 59.3 / \textbf{34.6} \\
        VGN ($\epsilon = 0.80$) & 85.8 / 89.5 & 59.8 / \textbf{51.1} & 84.0 / 79.9 & 78.4 / 69.0 & 52.8 / 30.1 \\
        \bottomrule
  \end{tabular}
  \label{table:grasp_results}
\end{table}

In order to evaluate the grasping performance of \gls{vgn}, we simulate a set of clutter removal benchmark scenarios. Each round, a new scene is generated with $m \in \{5, 10\}$ objects following the same ``pile'' or ``packed'' procedures used to generate the training dataset (see Section~\ref{sec:synthetic-training}). We test the performance using piles of simple geometric primitives (blocks), as well as piles and packed scenes with the 40 previously unseen test objects. The input data is generated by fusing images rendered from six viewpoints equally distributed along a circle placed \SI{50}{cm} above the workspace. We feed the reconstructed point cloud into \gls{vgn} and randomly select a grasp that has a predicted grasp quality higher than some threshold $\epsilon\in\{0.80,0.90,0.95\}$. The random selection is used to avoid repeatedly executing a bad grasp in case of a false positive, however this could also be resolved with some additional bookkeeping. Varying $\epsilon$ allows us to limit candidates to those with high predicted scores at the risk of a lower recall. Once a grasp is selected, we spawn the gripper at a pre-grasp pose, then linearly move along the approach vector until either the grasp pose is reached or a contact is detected. Next, we close the jaws of the gripper with constant force. The grasp is recorded as a success if the gripper can retreat without the object slipping out of the hand. One round of a clutter removal experiment consists of running the previous steps until (a) no objects remain, (b) the grasp planner fails to find a grasp hypothesis, or (c) two consecutive failed grasp attempts have occurred.

Performance is measured using the following three metrics averaged over 200/100 rounds with 5/10 objects, respectively:

\begin{itemize}
    \item \textbf{Success rate}: the ratio of successful grasp executions,
    \item \textbf{Percent cleared}: the percentage of objects removed during each round,
    \item \textbf{Planning time}: the time between receiving a voxel grid and returning a list of grasps.
\end{itemize}

For comparison, we also report results from repeating the same experiments with \gls{gpd}~\citep{gualtieri2016high}, an algorithm for synthesising 6 DOF grasps in clutter that only relies on a point cloud reconstruction of the scene. Since \gls{gpd} is stochastic by nature, we evaluate the highest ranked grasp with a positive score, meaning that the algorithm expects the grasp to be successful.

Table~\ref{table:grasp_results} shows success rates and percent cleared for the different experimental scenarios. In the experiments with 5 blocks, \gls{vgn} achieves high success rates while clearing significantly more objects compared to \gls{gpd}, especially in the packed scenes. We observe that decreasing $\epsilon$ results in lower grasp success rates, which shows that the network captured a meaningful estimate of grasp quality. The higher grasp success rates achieved with $\epsilon=0.95$ come at the cost of removing fewer objects from the scene, which is an expected trade-off. In general, we found that using a threshold of 0.9 leads to a good balance between success rate and percentage of cleared objects.

We also explored the performance in even denser scenes by running the same experiments with an object count of 10. Note that we only evaluate the ``pile'' scenario since it is unlikely for our ``packed'' scene generation procedure to place 10 objects in the workspace. We observe that both methods suffer from the increased complexity of the generated scenes. While the grasp success rates of~\gls{gpd} are impacted less, the algorithm seems more conservative removing only a fraction of the objects. We found the failure cases of VGN to be similar to the ones with smaller object numbers, consisting of a mix of small errors in position and orientation leading to objects slipping out of the hand, and occasional collisions between the gripper and its environment.

Where our method really shines is computation times. Inference of \gls{vgn} on a GPU-equipped workstation is really fast, requiring only \SI{10}{ms}, while \gls{gpd} on average required \SI{1.2}{s} to plan grasps on the same computer.

\subsection{Real Robot Experiments}

To validate that our model transfers to a real system with noisy sensing and imperfect calibration and control, we performed 10 rounds of table clearing experiments with a similar protocol to the simulation experiments. Each round, we randomly select 6 from the 12 test objects, place them in a box, shake the box to remove bias, then pour the contents of the box on a table in front of the robot. Each trial, the robot constructs the input to the grasp planner by continuously integrating the stream of incoming depth images along a pre-defined scan trajectory. For the grasp selection, we follow the approach of~\citet{gualtieri2016high} and execute the highest grasp candidate to avoid collisions between the robot arm and other objects in the scene. Grasps are recorded as a success if the robot successfully moves the object to a target bin located next to the workspace. We repeat the pipeline until either all objects are removed or two consecutive failed grasp attempts occur. We use the same trained network as in the simulated experiments with a grasp quality threshold $\epsilon$ of 0.9.

\begin{figure}[t]
    \centering
    \includegraphics[width=\textwidth]{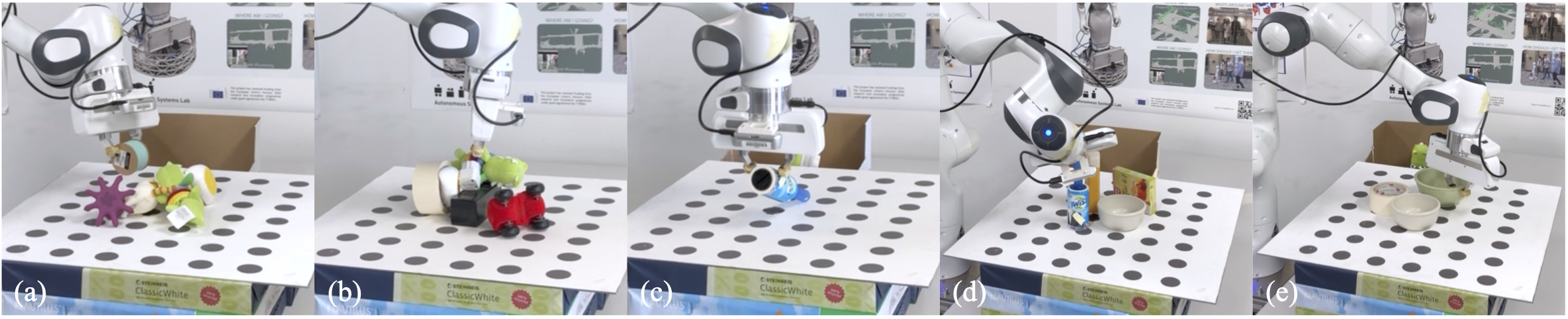}
    \caption{Examples of real world grasps detected by VGN (a)-(b). (c) shows a typical failure case for our model where the fingers slip off the cylinder-shaped object due to a small contact surface. The system is also capable of side-grasps (d) and picking the thin rim of bowls (e).}
    \label{fig:panda_grasps}
\end{figure}

Out of the 68 total grasp attempts, 55 were successful resulting in a success rate and percent cleared of 80\% and 92\%, respectively. Figure~\ref{fig:panda_grasps} shows some examples of successful and failed grasps. We observed that almost all failures (9/13) occurred while attempting a grasp on either the lint roller or hand gel. The reason for these failures was lack of friction when the fingers were not placed deep enough on these cylindrical objects as seen in Figure~\ref{fig:panda_grasps}(c). The root of this problem lies within the data generation procedure. Since the contact and friction models of the physics simulator used do not reflect the real world perfectly, configurations like the ones just described often result in a success in simulation, explaining the confident prediction of \gls{vgn}. The remaining failures were one collision with an object and three additional unstable grasps that resulted in the object slipping out of the hand. Two objects were pushed out of the workspace during grasp execution and therefore could not be moved to the target container. Since the on-board computer of the robotic platform does not have a GPU, we ran inference on the CPU only, taking on average \SI{1.25}{s}, significantly longer than using a dedicated graphics card.

We ran our grasping algorithm on 3 additional scenes consisting each of 4 packed, standing objects, including two bowls. There we showcase the flexibility of our system as the robot successfully performs a side-grasp on the lint roller and also manages to compute grasps on the thin edges of the bowls. In these three rounds, all objects were removed at first try except a box that was knocked over by the arm while moving a grasped object to the target container.

\section{Conclusion}

In this work, we presented our Volumetric Grasping Network, an end-to-end grasp synthesis approach that generates 6 DOF grasp proposals with a single forward pass of a Fully Convolutional Neural Network. We showed that we can train such a network on synthetic grasping trials to perform well in highly cluttered scenes without explicit collision checking. We also highlighted the efficiency of the approach when paired with a modern Graphics Processing Unit. We demonstrated that the model transfers directly to a real robotic setup without any additional adjustments thanks to the~\gls{tsdf}-based approach. Nevertheless, during the real robot experiments, we identified some limitations of training only on simulated data. It would be interesting to investigate whether introducing an adversarial test to the physics simulation similar to~\citep{pinto2017supervision} would allow us to overcome this issue. Also, while the presented approach can be easily scaled to different gripper geometries, this needs to be verified with further experiments. Finally, in a next step we would like to build on top of the real-time capabilities by closing the visual feedback loop, either to react to dynamic changes or to re-plan the grasp execution online instead of relying on a fixed scan trajectory.



\clearpage
\acknowledgments{This work was funded in part by ABB Corporate Research, the Amazon Research Awards program, and the Luxembourg National Research Fund (FNR) 12571953.}


\bibliography{bibliography}

\end{document}